\documentclass[]{interact}

\usepackage{epstopdf}
\usepackage[caption=false]{subfig}

\usepackage[numbers,sort&compress]{natbib}
\usepackage{amsmath, amssymb}

\usepackage{todonotes}
\usepackage[ruled,linesnumbered]{algorithm2e}
\usepackage{multirow}
\usepackage{multicol}
\usepackage{tabularx}

\bibpunct[, ]{[}{]}{,}{n}{,}{,}
\renewcommand\bibfont{\fontsize{10}{12}\selectfont}
\makeatletter
\def\NAT@def@citea{\def\@citea{\NAT@separator}}
\makeatother

\theoremstyle{plain}

\theoremstyle{definition}

\theoremstyle{remark}

\begin{document}

\articletype{full paper}

\title{Multimodal Grounding for Embodied AI via Augmented Reality Headsets for Natural Language Driven Task Planning.}


\author{
\name{Selma Wanna \textsuperscript{a}\thanks{CONTACT Selma Wanna Email: slwanna@utexas.edu}, Fabian Parra\textsuperscript{a}, Robert Valner\textsuperscript{b}, Karl Kruusamäe\textsuperscript{b} and Mitch Pryor\textsuperscript{a}}
\affil{\textsuperscript{a}University of Texas at Austin; \textsuperscript{b}University of Tartu}
}

\maketitle

\begin{abstract}
Recent advances in generative modeling have spurred a resurgence in the field of Embodied Artificial Intelligence (EAI). EAI systems typically deploy large language models to physical systems capable of interacting with their environment. In our exploration of EAI for industrial domains, we successfully demonstrate the feasibility of co-located, human-robot teaming. Specifically, we construct an experiment where an Augmented Reality (AR) headset mediates information exchange between an EAI agent and human operator for a variety of inspection tasks. To our knowledge the use of an AR headset for multimodal grounding and the application of EAI to industrial tasks are novel contributions within Embodied AI research. In addition, we highlight potential pitfalls in EAI's construction by providing quantitative and qualitative analysis on prompt robustness.

\end{abstract}


\section{Introduction}
Offloading dangerous inspection, surveillance, and manipulation tasks to robots in unstructured environments, e.g., industrial task domains, incident response, etc. has been the driving motivation for utilizing robots in human-robot teams. However, the supervision of such a team requires the human operator to work and lead the robots simultaneously. To do so effectively requires an intuitive and minimally restrictive control interface that also provides sufficient situational awareness of the robots and the environment. Mixed Reality (MR) technology offers a capable platform for designing such control interfaces, as allows overlaying the operator’s view with, e.g.,  heat-maps, structural weak-spots, and the location of other human and robot team-members in no line-of-sight or low visibility. 

\begin{figure*}[ht!]
    \centering
    \includegraphics[width=0.6\textwidth]{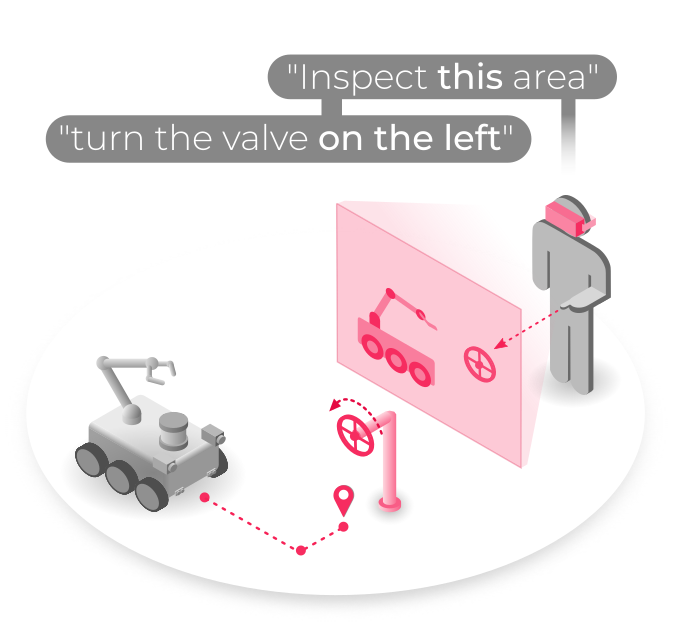}
    \caption{Conceptual overview of mixed reality headset utilized as an intuitive and minimally restrictive multimodal control interface.}
    \label{fig:graphical_abstract}
\end{figure*}

MR tools, such as Augmented Reality (AR) headsets often come equipped with hand-tracking and speech recognition capabilities, allowing the operator to utilize multiple naturalistic communication modalities that can reduce ambiguity of unimodal interaction in HRI (Fig. \ref{fig:graphical_abstract}). The challenge, however, is understanding the operator’s intent from the combination of these modalities. Commands issued via speech and gestures have to be robustly grounded into executable tasks.

We address this issue of multimodal grounding by adapting prior work in Embodied AI research: the study of artificial intelligence deployed to physical systems capable of interacting with their environments \cite{IEEE_embodiedai_survey, embodied_ai_retrospective_survey} to AR headsets. Specifically, we inject visual and language information obtained via the AR headset directly into the language prompt of GPT-3 \cite{GPT3_paper}. However, to extend EAI to generalize to industrial settings, we enforced a co-located, human-robot teaming paradigm where an AR headset mediated dialogue between the EAI agent and human operator. 

To our knowledge, this demonstration is novel to EAI deployments with respect to the industrial domain and the use of the AR headset. We contribute a successful demonstration of EAI and AR integration, provide studies on prompt design which highlight potential pitfalls in common EAI constructions \cite{singh2022progprompt, Huang2022EAI, saycan, zeng2022socratic, probes-liang-etal-2022-visual, vemprala2023chatgpt} with respect to prompt fragility, and conclude with a holistic discussion on the merits of adopting EAI for multimodal task planning.

\section{Related Work}
Prior designs for EAI deployments have largely converged on a common architecture which leverages hard prompt learning techniques in conjunction with object detectors to generate an agent's next action prediction. A summarization of prior work is provided in Table \ref{tab:robot_prompting_research}.

\begin{table}[ht]
\centering
\resizebox{\textwidth}{!}{\begin{tabular}{ |c|c|c|c|c|c|c|}

\hline
Method, Year \& Reference & Prompt Type & Model(s) & Tasks\\
\hline
\hline
ProgPrompt, 2022 \cite{singh2022progprompt} & Hard & GPT-3 \cite{GPT3_paper} & VirtualHome \cite{Puig_2018_CVPR} + Physical  \\ 
\hline
InnerMonologue, 2022 ~\cite{Huangzeroshotplanner2022,Huang2022EAI} & Hard  & PaLM \cite{PALM} + InstructGPT \cite{InstructGPT} & Physical \\
\hline
SayCan, 2022 \cite{saycan} & Hard  & PaLM \cite{PALM} & ALFRED \cite{ALFRED20} + BEHAVIOR \cite{li2022behaviork} + Physical  \\
\hline
SocraticModels, 2022 \cite{zeng2022socratic} & Hard  & ViLD \cite{gu2022openvocabulary}  & Simulated Tabletop  \\
\hline
ProbES, 2022 \cite{liang-etal-2022-visual}  & Soft & ViLBERT \cite{vilbert} with LSTM and MLP Head & REVERIE \cite{Qi_2020_CVPR} + R2R \cite{mattersim}\\ 
\hline
\end{tabular}
}
\caption{\label{tab:robot_prompting_research} Summarization of recent effort to leverage LLM prompting techniques for Embodied AI.
}
\end{table}

Most methods that rely on hard prompt engineering employ human-engineered prompts with little quantitative design justification. This is problematic because prompt engineering is a non-robust process where different but semantically equivalent prompts may cause task performance to vary between pure chance and state-of-the-art performance \cite{lu-etal-2022-fantastically}. The fickleness of prompt design manifests itself in non-robust prompt preference \cite{cao-etal-2022-prompt}, prompt sample selection bias \cite{lu-etal-2022-fantastically}, and sample ordering bias \cite{lu-etal-2022-fantastically}. Additionally, discrete prompt design often requires extensive human-engineering \cite{prompttuning_survey} thus leading to the recent creation of PromptCraft, an open-source platform for robotics researchers to share their prompting strategies \cite{vemprala2023chatgpt}. 

The emergence of multimodal foundation models \cite{FoundationModel} such as CLIP \cite{CLIP} have reinforced the reliance on LLMs for Embodied AI task planning \cite{Khandelwal_2022_CVPR_CLIP_embodiedAI, ThomasonClip2022, liang-etal-2022-visual}. These methods, which distill image information into text, are commonly leveraged for tasks such as object detection \cite{gu2022openvocabulary} and scene description \cite{kamath2021mdetr} as a means of world and agent state tracking. These systems are uniquely compatible with the LLM prompting paradigm because the LLM is restricted to text modalities by design. Within this LLM-driven design, there are largely two multimodal fusion techniques: injecting visual information via image-to-text algorithms into the language prompt \cite{singh2022progprompt, Huang2022EAI, Huangzeroshotplanner2022, zeng2022socratic} or synthesizing the information downstream \cite{probes-liang-etal-2022-visual, saycan}. We adopt the former approach; however, we abandon image-to-text models in favor of human generated virtual reality (VR) markers.

\section{Background}

\subsection{Large Language Models}
Large Language Models (LLMs) come in a variety of sizes and architectures; however, this discussion centers on autoregressive models capable of in-context learning \cite{GPT3_paper} because they are best suited for text generation tasks \cite{prompttuning_survey}, e.g., task planning \cite{saycan, singh2022progprompt, Huang2022EAI}. For autoregressive language models, the most common training objective is to maximize the log-likelihood of the next token prediction at a decoding step, \(t\), based on the context provided by the previous \(t-1\) tokens. This is formalized in Equation \ref{eq:llm_obj} where, \(\mathbf{y}\), represents the decoded text that is generated as a result of conditioning on the input text, \(\mathbf{x}\), and latent features, \(h\). Equation \ref{eq:llm_obj} aims to solve for the LLM parameters, \(\theta\), that maximize the log likelihood of the observations, \(y\).

\begin{equation}
    \max_{\theta}\ log\ p(\mathbf{y}|\mathbf{x}; \theta) = \max_{\theta} \sum_{y_t} log\ p(y_t | h_{<t};\theta)
    \label{eq:llm_obj}
\end{equation}

This training objective is performed under self-supervision tasks \cite{FoundationModel}. Under this construction, the model learns language information by solving a variety of de-noising tasks such as masked token prediction \cite{BERT}, next sentence prediction \cite{BERT}, next token prediction (language modeling) \cite{GPT3_paper}, long range dependency modeling \cite{paperno-etal-2016-lambada}, etc. For a comprehensive overiew of de-noising objectives and functions, please refer to \cite{prompttuning_survey}.

\subsection{Prompt Learning} 
\label{greedysearch}
The discovery of in-context learning \cite{GPT3_paper} in conjunction with the expensiveness of training enormous language models drove the field of natural language research toward prompt based learning \cite{FoundationModel}. Additionally, the companies that offer LLMs as a service restrict access to LLM feature and gradient information, rendering other transfer learning techniques, such as fine tuning, infeasible \cite{ye2022durrett}. As such, this work focuses on tuning-free prompt learning: a type of prompt engineering that involves searching for the optimal prompting function for a LLM with frozen model parameters \cite{prompttuning_survey}.

Formally, prompt learning involves taking natural language, \(\mathbf{x}\), as an input to a prompt function, \(f_{prompt}(\cdot)\), to generate a prompt: \(\mathbf{x}'\). While the theory behind prompt learning lags behind its empirical findings, it is speculated that prompting incantations \cite{HELM} can prime \cite{GPT3_paper} LLMs to activate relevant neurons for a desired task \cite{su-etal-2022-transferability}.

The simplest method for prompt searching is to perform an exhaustive search along the axes of few-shot example ordering, example selection, and number of examples. This search space is limited by the maximum token request and rate limits API constraints provided by OpenAI's \texttt{text-davinci-003} model \cite{GPT3_paper}. As such, the search space comprised of the set of 2-length permutations of multimodal UMRF decoding examples. Each example permutation was scored by its BLEU score \cite{papineni-etal-2002-bleu} accuracy against its natural language instruction's ground-truth UMRF decoding in the validation set. The prompt permutation with the highest accuracy was chosen as the optimal prompt.

\subsection{Unified Meaning Representation Format}

\begin{figure*}[ht!]
    \centering
    \includegraphics[width=1.0\textwidth]{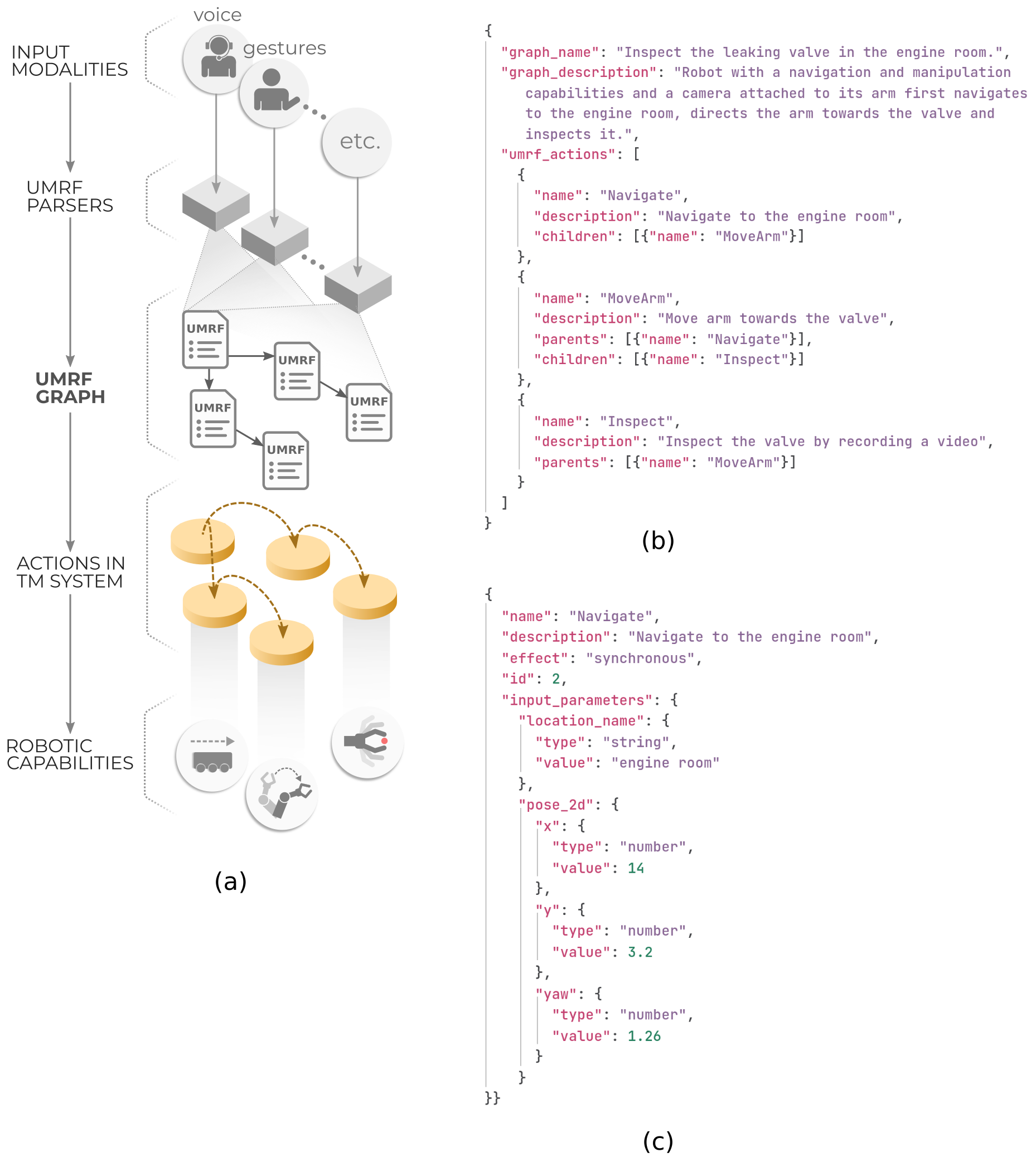}
    \caption{(a) UMRF graph based robot commanding pipeline where arbitrary command modality, e.g., voice command, is parsed and converted to UMRF graph notation via a dedicated parser. Each UMRF node in the UMRF graph is then mapped to a known executable action that implements the desired behaviour. (b) Example of an UMRF graph in JSON notation outlining an inspection task that contains three sequential actions. (c) A detailed UMRF JSON notation of a parametrized navigation action that accepts coordinates and the name of the target location.}
    \label{fig:umrf_pipeline}
\end{figure*}

The Unified Meaning Representation Format (UMRF) \cite{UMRF} is a platform independent task description format based on JSON notation, designed to decouple robot's autonomous capabilities and its command interface (Fig. \ref{fig:umrf_pipeline}a). Thus a robot could be controlled via any system or command interface that outputs commands in UMRF, increasing both the robot's and command interface's modularity and reusability. Tasks are defined as graphs of interconnected actions, described via parent/child relations (Fig. \ref{fig:umrf_pipeline}b). UMRF supports sequential, concurrent and cyclical graphs and parametrization, i.e., actions can accept and produce data (Fig. \ref{fig:umrf_pipeline}c). A thorough coverage of UMRF can be found in \cite{UMRF}.

\section{UMRF Prompt Design}
\label{prompt_design}
\begin{figure}[h]
    \centering
    \includegraphics[scale=0.6]{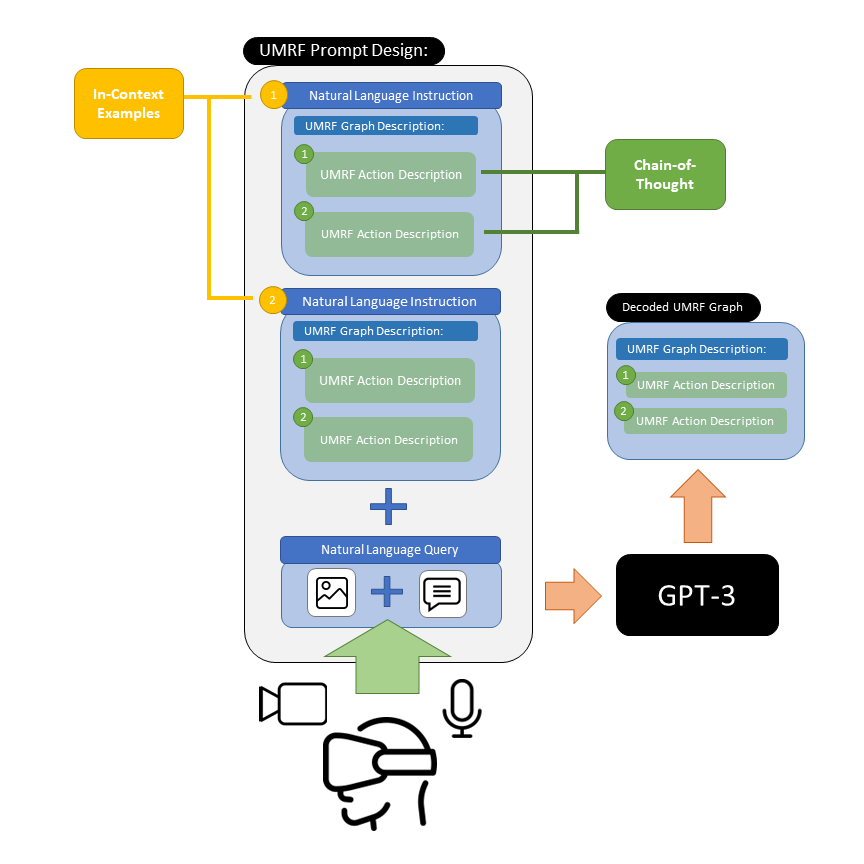}
    \caption{Summarization of the UMRF prompt design with multimodal input and output streams labeled.}
    \label{fig:prompt_design}
\end{figure}

Our discrete prompt methodology is most similar to InnerMonologue \cite{Huang2022EAI} where we incorporate relevant world contexts and objects in conjunction with few shot examples within the prompt. However, we depart from their effort by (1) incorporating chain-of-thought prompting techniques within in-context examples \cite{chain-of-thought}, similar to ProgPrompt \cite{singh2022progprompt}; (2) adopt the UMRF \cite{UMRF} formalism for action decoding instead of pythonic code generations; and lastly (3) provide language and visual feedback via the AR headset to develop a multimodal prompt. For greater detail, please refer for Figure \ref{fig:prompt_design}.


\section{Prompt Design Experiments}
\subsection{Experiment 1. Greedy \& Exhaustive Search}

The exhaustive search algorithm explored a space of \(_{10}P_{2}\) prompts evaluated on a validation set of five examples. To emphasize the expense of the greedy search, the average query time to OpenAI was 1.5 minutes. Thus, running a search of over 450 queries cost roughly 11 hours of computation. This expense motivates the need for more efficient search algorithms for prompt design such as the method suggested in Section \ref{experiment2}. 

\begin{figure}[h]
    \centering
    \includegraphics[scale=0.35]{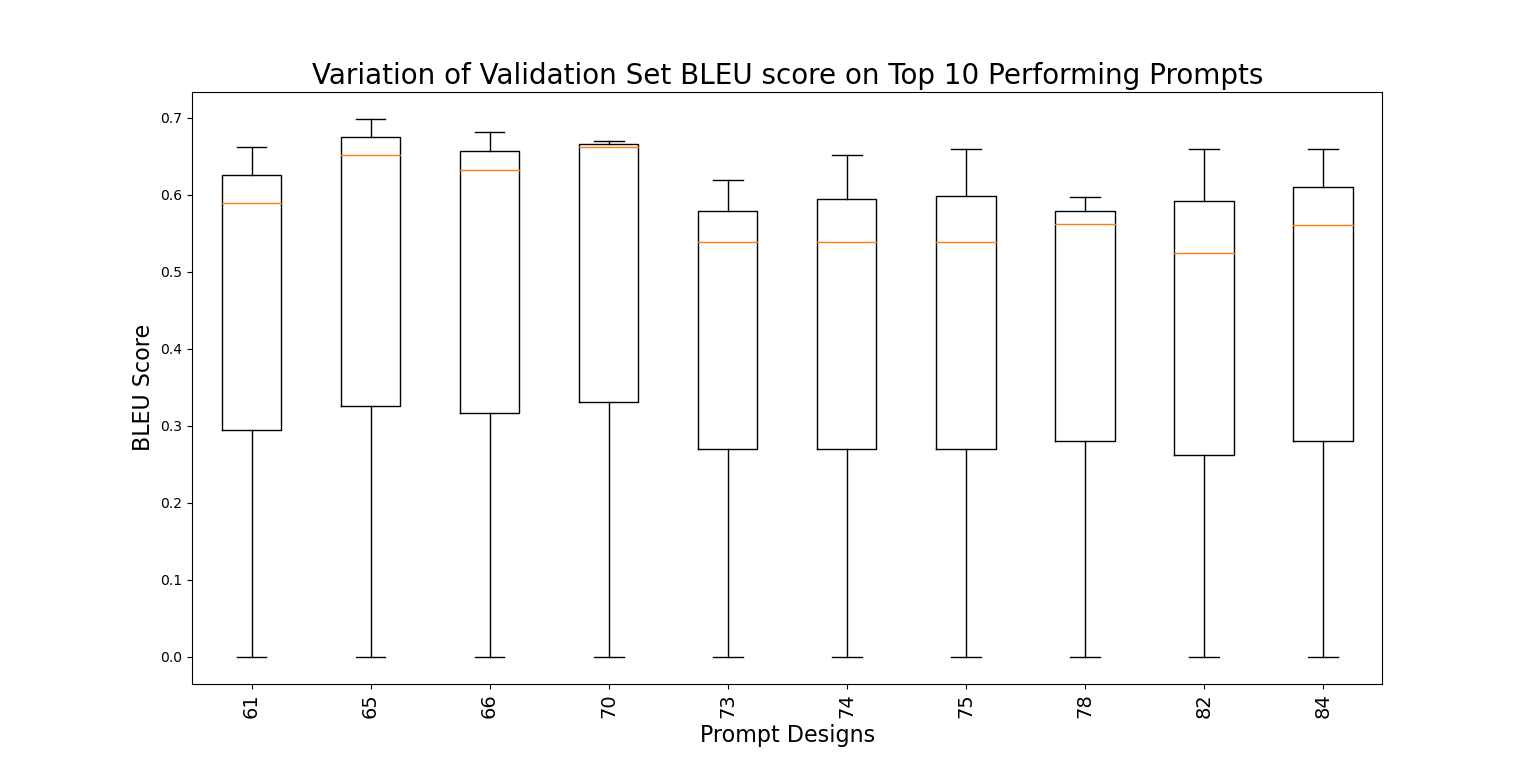}
    \caption{The variation in performance of the top 10 performing prompt designs. Note that despite being stronger prompts overall, there is a lack of robust generalization for each design to the validation set.}
    \label{fig:top_10_prompting_var}
\end{figure}

The experiment outcomes support our hypothesis that prompt design is highly non-robust even for simple tasks that can be solved using traditional grammars or simpler neural networks, such as the UMRF decomposition task. This is clearly demonstrated in appendix \ref{fig:prompting_var}. The top ten performing prompts are provided in Figure \ref{fig:top_10_prompting_var}.

While information ordering and example selection account for high variation (see Figures \ref{fig:prompt_info_ordering} and \ref{fig:prompting_ex_importance} respectively), the top performing prompts most commonly share combinations of example types: 1, 4, and 5 as described in Table \ref{tab: example_types}. In every case, barring one, example type 5 performed best as the first example in the sequence. This may be because example 5 is both the longest and most informative training example. While its complexity may allow for the best generalization performance when ordered first, its placement toward the end of the prompt may confuse the LLM as it begins to decode the validation query. 

The strongest performing prompt design with an average BLEU score of 0.662 was prompt 70 with the structure: example 4 + example 5 (please refer to Table \ref{tab: top_10_struct}.) The prompt with the highest BLEU score (0.850) had the following structure: example 1 + example 4. For more details regarding the prompt design of the top ten prompts, please refer to Table \ref{tab: top_10_struct}.

\subsection{Experiment 2. Assessing Prompt Fragility}
\label{experiment2}
In an attempt to better explore the prompt search space, we extended prior work in textual data augmentation \cite{ren2021taa} to find an optimal discrete prompt. Specifically, we tried to learn a prompt policy by searching for compositional augmentations which maximize BLEU score \cite{papineni-etal-2002-bleu} on a validation set through Bayesian optimization. The value in this methodology is in learning a surrogate model that can more cheaply mimic the outputs of an LLM, allowing for more efficient search space exploration. However our adaptation of Text AutoAugment \cite{ren2021taa} to generative language tasks was unsuccessful due to weak reward signals and high performance sensitivity to heuristic augmentations as discussed in \cite{ren2021taa}. 

\begin{figure}[h]
    \centering
    \includegraphics[scale=0.65]{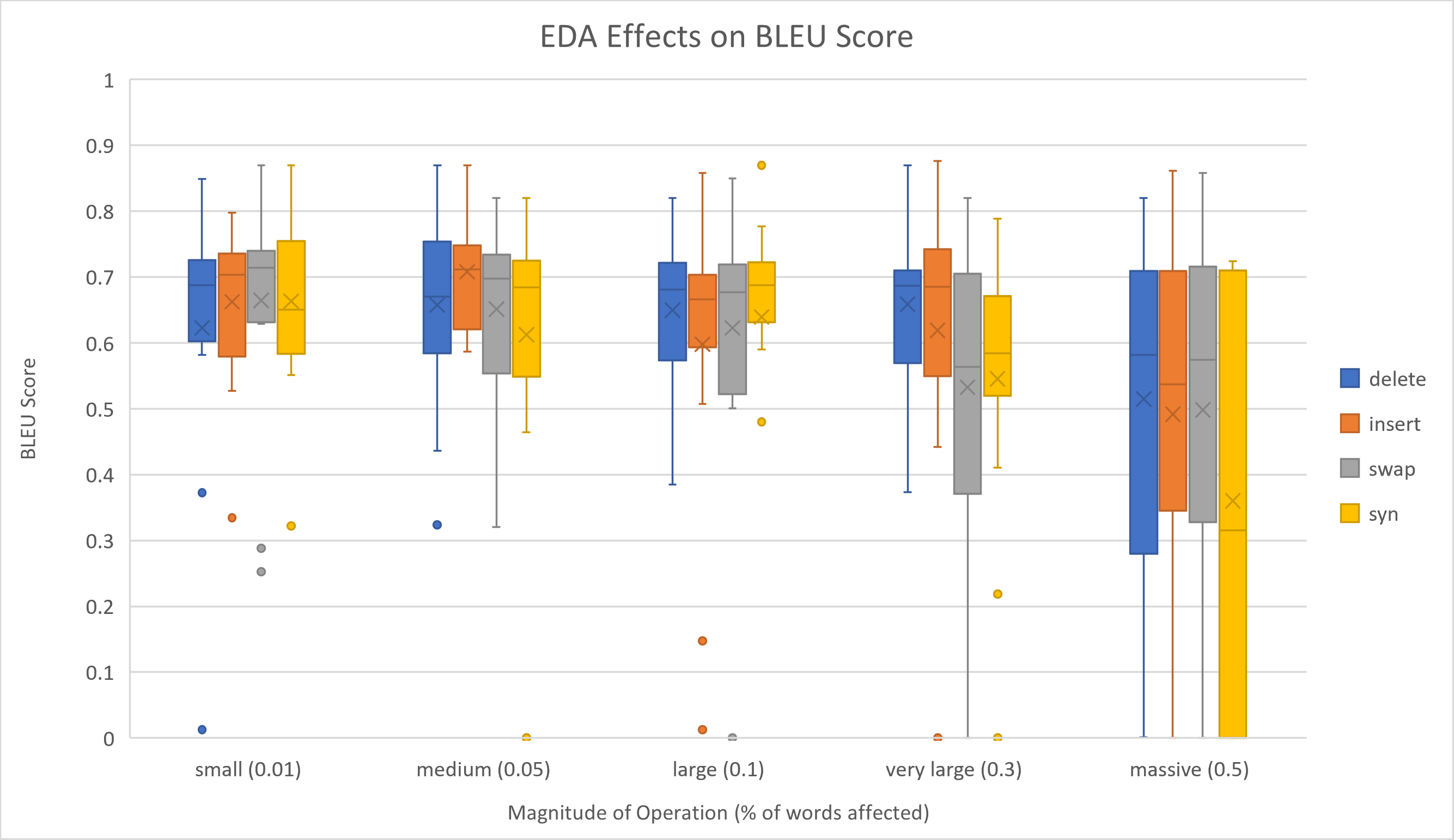}
    \caption{Prompt sensitivity analysis to EDA \cite{wei-zou-2019-eda} perturbations at various magnitudes.}
    \label{fig:eda_robustness_1}
\end{figure}

Additional experiments regarding prompt fragility were conducted to measure BLEU score accuracy given a single application of text augmentation with varied magnitudes of the operation \cite{wei-zou-2019-eda}. As shown in Figure \ref{fig:eda_robustness_1}, GPT-3 is robust to random deletion and insertion operations. However, random swapping of words and synonym replacements tend to have a larger effect on performance.

\begin{figure}[h]
    \centering
    \includegraphics[scale=0.50]{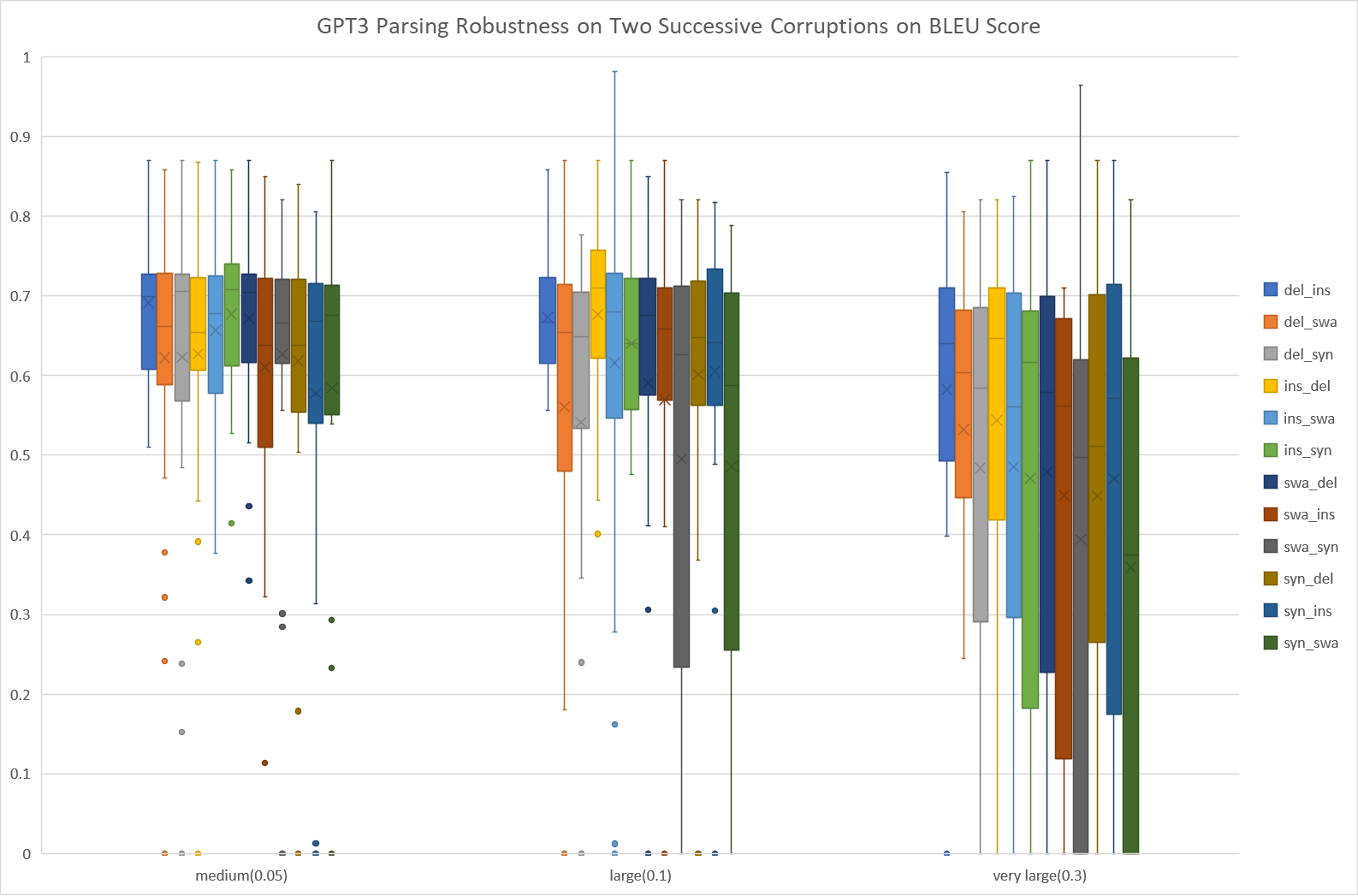}
    \caption{Prompt sensitivity analysis to EDA \cite{wei-zou-2019-eda} after compositional perturbations were performed at various magnitudes.}
    \label{fig:eda_robustness_2}
\end{figure}

A more representative experiment was performed to measure the BLEU score accuracy after applying compositional augmentations to the prompt. The results are shown in Figure \ref{fig:eda_robustness_2}. Generally, compositional augmentations widened performance variation across the board. As a future recommendation, the search space for the magnitude parameter for Text AutoAugment \cite{ren2021taa} should be constrained to less than 0.1.

\begin{figure}[h]
    \centering
    \includegraphics[scale=0.65]{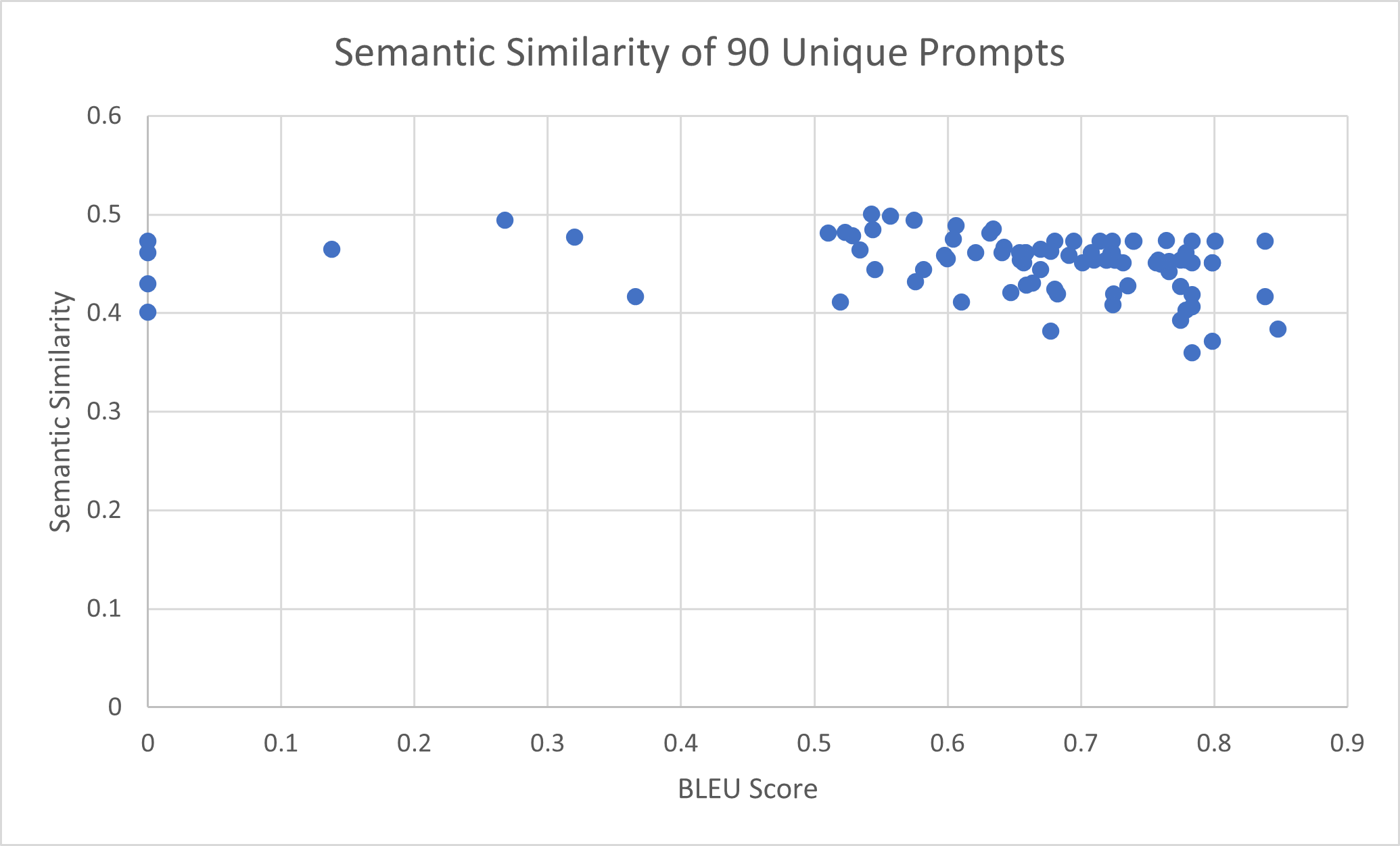}
    \caption{The semantic similarity of the 90 unique prompts generated using the greedy search algorithm from Section \ref{greedysearch} measured against UMRF examples seen in GPT-3's pretraining dataset \cite{thepile}.}
    \label{fig:sem_sig_meth1}
\end{figure}

\begin{figure}[h]
    \centering
    \includegraphics[scale=0.5]{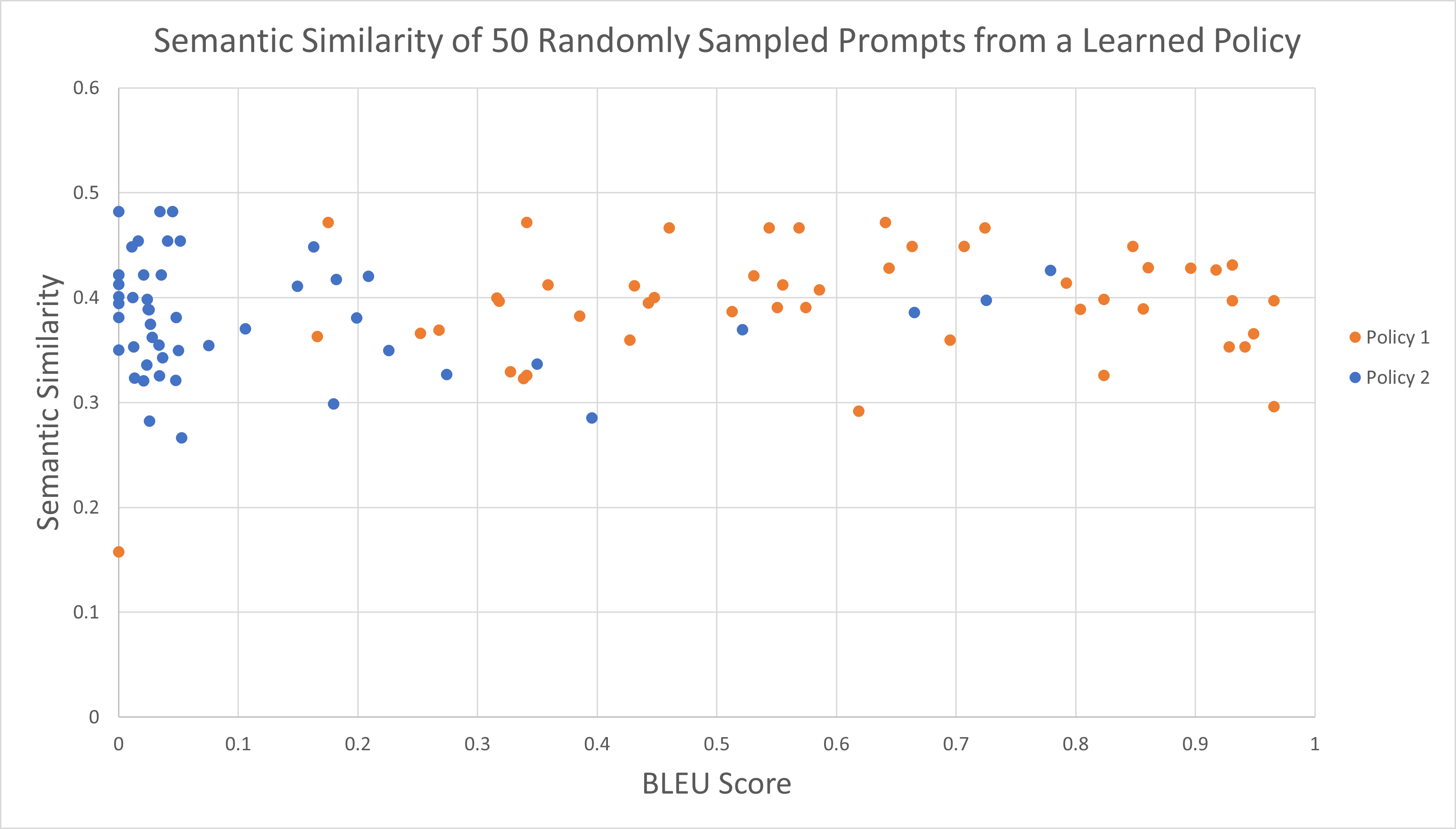}
    \caption{The semantic similarity of prompts sampled from our adaptation of Text Autoaugment  against UMRF examples seen in GPT-3's pretraining dataset \cite{thepile}. Policy 1 did not apply heuristic data augmentation techniques \cite{wei-zou-2019-eda} while policy 2 did through application of synonym replacement.}
    \label{fig:sem_sig_meth2}
\end{figure}

Despite this setback, further analyses was conducted to investigate whether a given prompt's similarity to UMRF examples present in GPT-3's pretraining dataset could be a predictor of a prompt's performance. Semantic similarity was measured as the cosine similarity between \texttt{all-MiniLM-L6-v2} model embeddings \cite{reimers-2019-sentence-bert} of a prompt versus the UMRF examples present in The Pile \cite{thepile}. In our low-data regime, we could not identify a correlation (see Figures \ref{fig:sem_sig_meth1} and \ref{fig:sem_sig_meth2}).

From this analysis we are able to identify a legitimate danger in carelessly deploying EAI in safety-critical environments. Specifically, semantically equivalent prompts can vary greatly in performance. By qualitative inspection, seemingly harmless applications of synonym replacement such as converting numerical representations to their written forms as well as replacing similar words, e.g., `move forward' to `approach' and `table' to `bureau', noticeably harmed task performance. Potentially adversarial augmentations, including converting the coordinate variable, `y', to `yttrium' or `atomic number 39' were detrimental to task performance. Unfortunately, there is no immediate path forward for roboticists developing systems in niche domains. Specifically, we cannot rely on robustness techniques employed for language tasks with greater task representation in LLM pretraining datasets \cite{arora2023ask}.

\section{Demonstration}

\begin{figure*}[ht!]
    \centering
    \includegraphics[width=0.8\textwidth]{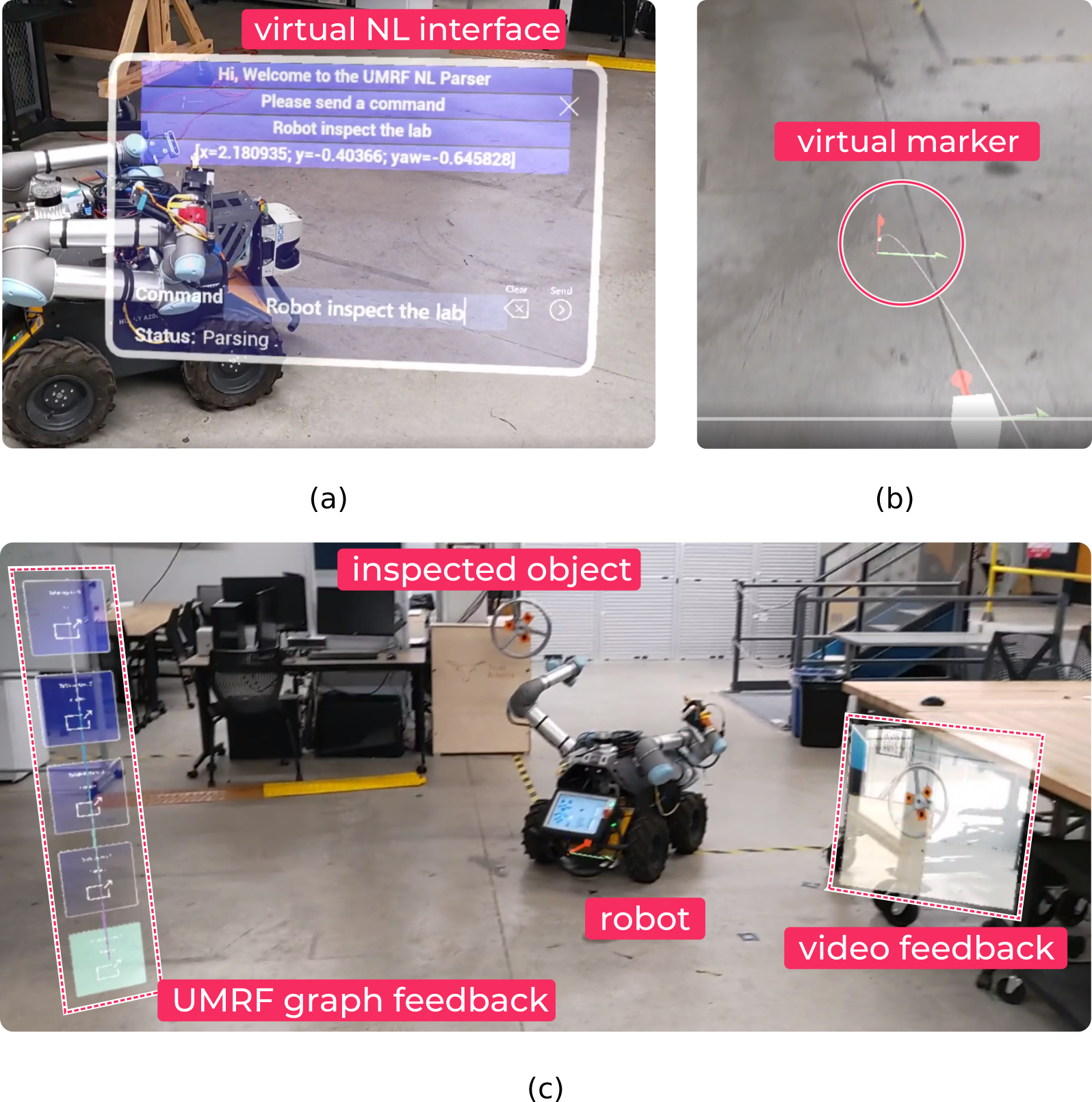}
    \caption{AR interface from the operator's perspective. (a) interactive marker that user can drag \& drop (b) Natural Language interface to send voice commands. (c) UMRF graph and video feedback is shown in the AR space}
    \label{fig:demo_interface}
\end{figure*}

This section demonstrates the functionality of the multimodal speech and augmented reality based UMRF graph parser, outlined in Section \ref{prompt_design}. The demonstration depicts a remote inspection scenario, where a mobile manipulator robot (Clearpath Husky + two Universal Robots UR5's), equipped with a camera (Intel RealSense D435) attached to the end-effector, has to navigate to and inspect specific areas defined by the operator. The operator is equipped with an AR headset (Microsoft HoloLens 2) that is able to capture voice commands (Fig. \ref{fig:demo_interface}a) and allows defining goal locations via gesture-operated virtual markers (Fig. \ref{fig:demo_interface}b). Inspection and task execution feedback is overlaid to the operator's field of view in real-time (Fig. \ref{fig:demo_interface}c). The Azure Spatial Anchors plugin \cite{spatialanchors} was used on HoloLens to allow the robot and AR-devices to co-localize and share same the same reference frame. Target poses are generated by spawning a coordinate frame in the world, and dragging it to the desired pose. The Natural language command is captured by pressing the microphone icon on the HoloLens app.

\begin{figure*}[ht!]
    \centering
    \includegraphics[width=0.95\textwidth]{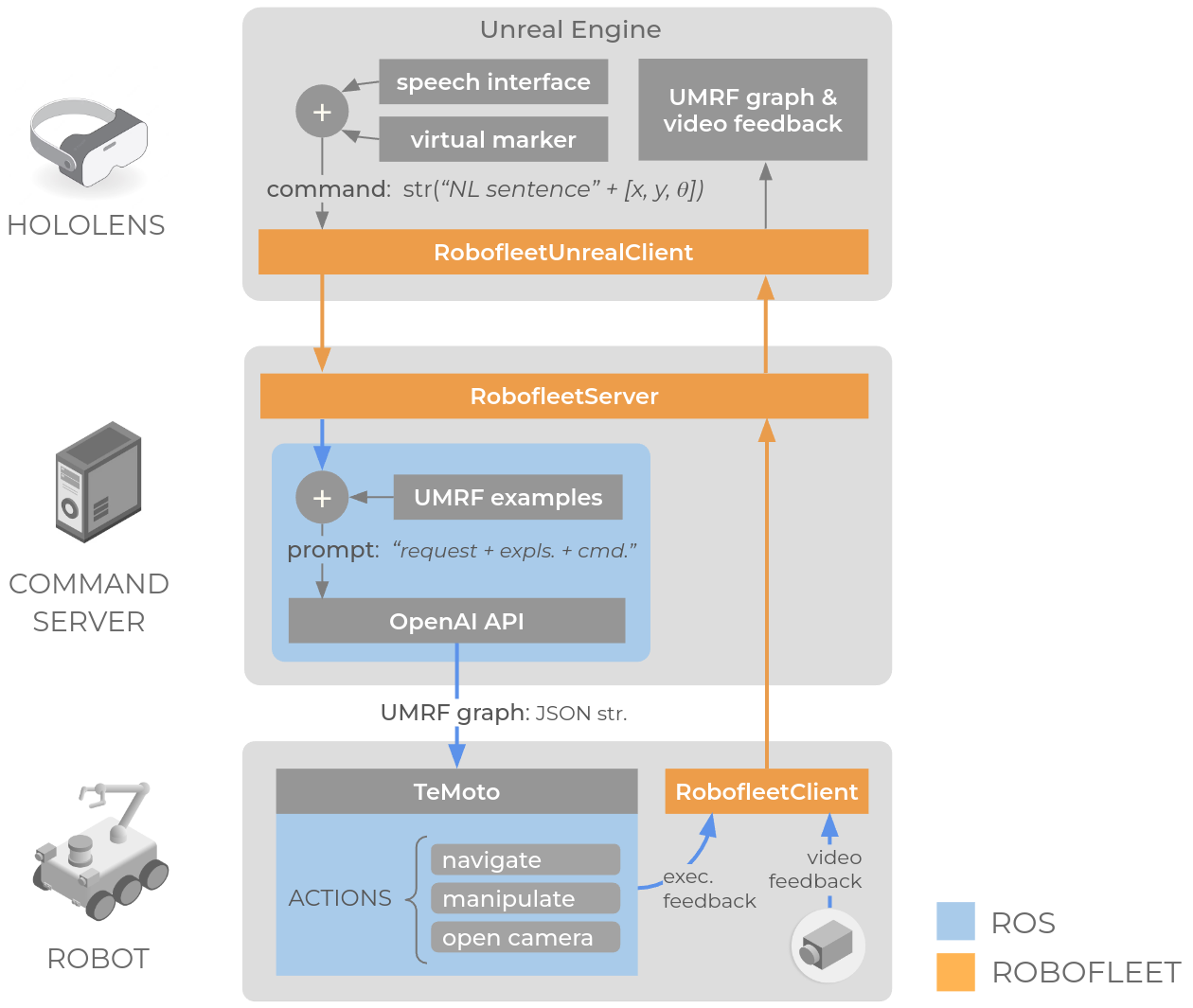}
    \caption{Software setup of the demonstration}
    \label{fig:demo_setup}
\end{figure*}

Fig. \ref{fig:demo_setup} shows the software setup of the demo, containing three main components: HoloLens, which captures the operator's input and provides feedback; command server, which hosts the UMRF parser; and the robot, that is able to execute tasks outlined in UMRF notation. The Robot Operating System (ROS) \cite{ros} and RoboFleet \cite{sikand2021robofleet} were used for data distribution between the components. The operator interface on HoloLens2 was implemented in Unreal Engine 4.26 \cite{unrealengine}, which combines an operator's voice command with the coordinates of the virtual marker to a string format. The combined input is sent to the command server via RoboFleetUnrealClient. The UMRF parser (available on GitHub\footnote{github.com/temoto-framework/gpt\_umrf\_parser}) on the command server, implemented as a ROS Python node, receives the command string and constructs the prompt (see Section \ref{prompt_design}). Each prompt embeds five \texttt{operator command + UMRF graph} pair examples (Table \ref{tab:umrf_examples}), which then is sent to OpenAI via openai v.0.25.0 Python API. The API call was configured for \texttt{text-davinci-003} model, with \texttt{max\_tokens=1024}, and rest of the settings on default values. The UMRF JSON string, retured by OpenAI, then is sent to the robot via ROS message. TeMoto Framework \cite{temoto} was used to both control the mobile base, camera, and manipulators of the robot, as well as TeMoto Action Engine was used to ground the UMRF graphs to executable actions (setup files available on GitHub\footnote{github.com/temoto-framework-demos/gpt\_temoto\_demo}).

\begin{table*}
  \caption{Examples used for constructing the prompts. Only abstract description of the output is provided, all examples in full detail available in online materials.}
  \label{tab:umrf_examples}
  \small
  \begin{tabularx}{\textwidth}{p{1mm} p{130mm} }
    \toprule
      & Examples \\
    \midrule
    1 & INPUT: \textit{`Move to the main hall [x=14; y=3.2; yaw=1.26]'} \\
      & OUTPUT: single `navigation' action \\
      & \\
    2 & INPUT: \textit{`Go to the workshop [x=-33.9; y=12.1; yaw=0.04]'} \\
      & OUTPUT: single `navigation' action \\
      & \\
    3 & INPUT: \textit{`robot go observe the valve [x=-93.6; y=11.0; yaw=-0.85]'} \\
      & OUTPUT: sequence of: $\text{`navigate'}\rightarrow\text{`manipulate'}\rightarrow\text{`scan'}\rightarrow\text{`manipulate'}\rightarrow\text{`scan'}$ \\
      & \\
    4 & INPUT: \textit{`robot go inspect the workshop [x=74.2; y=-223.6; yaw=2.72]'} \\
       & OUTPUT: sequence of: $\text{`navigate'}\rightarrow\text{`manipulate'}\rightarrow\text{`scan'}\rightarrow\text{`manipulate'}\rightarrow\text{`scan'}$ \\
      & \\
    5 & INPUT: \textit{`Scan the area'} \\
      & OUTPUT: single `scan' action \\
      & \\
  \bottomrule
  \end{tabularx}
\end{table*}

\section{Discussion \& Future Work}
In this paper we provide successful demonstrations of inspection tasks in industrial settings by mediating multimodal information through an AR headset. Despite the efficacy of the EAI construction, the prompting paradigm necessitates greater scrutiny from robotics researchers. Specifically, prompt designs that leverage in-context learning are not token-space efficient. This prevents LLMs from observing larger quantities of training examples within the hard prompt construction. Additionally, extensive human-engineering is required to develop `optimal' discrete prompts. This is partially due to prompt fragility. Furthermore, the nonrobustness of prompts to natural and adversarial perturbations add on additional vulnerabilities to physical systems. Lastly, there are technical challenges when relying on a third party to serve a LLM, particularly during periods of high demand. Commonly, OpenAI API request and rate limits were exceeded and led to no API responses or incomplete parses.

We outline avenues of future work as follows. First, there is a need to conduct additional studies on  human operator agency and quality-of-life in our collaborative human-robot team setup versus the traditional human-in-the-loop paradigm where the operator's role is relegated to correcting erroneous vision algorithm outputs. Second, it is necessary to quantify the gap in EAI performance when using human-assisted AR markers versus imperfect object detectors. Third, the robot dialogue responses which indicate agent and world state information are entirely visual. Efforts to implement text-to-speech algorithms may improve operator enjoyment of using the system. Fourth, to address the issue of token space efficiency, exploring multi-step LLM prompting techniques is necessary. A potential path forward is to first query for a compact representation of the task graph then query the LLM to fill in the nodal information. Fifth, it is necessary to conduct a study on task performance and robustness of various representations for LLM task planners to decode natural language into, e.g., UMRFs, pythonic code \cite{singh2022progprompt} or lists \cite{Huang2022EAI} as a function of their representation frequency in GPT-3's pretraining dataset. We speculate more representative formalisms may allow the use of information retrieval solutions to the prompt-robustness challenge \cite{arora2023ask}. Lastly, given the preliminary outcomes on prompt fragility, it is imperative that a sensitivity analysis be conducted between the magnitude of prompt perturbations and their effects on validation BLEU scores. This is our immediate next step toward characterizing prompt robustness for EAI systems.

\section*{Acknowledgement}

We thank Los Alamos National Laboratory for their support. LA-UR-23-22632.

\section*{Funding}

This research has been in part supported by European Social Fund via IT Academy programme, Estonian Centre of Excellence in IT (EXCITE) funded by the European Regional Development Fund, and AI \& Robotics Estonia co-funded by the EU and Ministry of Economic Affairs and Communications in Estonia.

Additionally, we thank the support of Los Alamos National Laboratory and the Center for Nonlinear Studies.


\section*{Notes on contributors}


\textbf{ Selma Wanna} received a B.Sc. degree in electrical engineering and M.Sc. degree in mechanical engineering from the University of Texas at Austin. Her research interests include quantifying the robustness of deep learning systems operating in out-of-distribution domains.

\textbf{Fabian Parra} received his bachelor’s degree in mechatronic engineering from the Nueva Granada Military University in Bogota, and the M.Sc. degree in Robotics and Computer engineering from the University of Tartu in Estonia. His research interests include mobile manipulator robots, supervised autonomy and human-machine interfaces.

\textbf{Robert Valner} is a junior researcher and a Ph.D. candidate at University of Tartu. He received B.Sc. and M.Sc. degrees in physics and computer engineering respectively from University of Tartu. His current research interests include fault tolerant and adaptive robotic architectures, multi-robot systems, human-robot interaction and autonomous robotic systems.

\textbf{Dr. Karl Kruusamäe} is an associate professor of robotics engineering at the University of Tartu. He received the M.S. degree in information technology and the Ph.D. in physics from the University of Tartu, Tartu, Estonia, in 2008 and 2012, respectively. His research interests include human-robot interaction and shared autonomy.

\textbf{Dr. Mitch Pryor} is a Senior Research Scientist and Lecturer for the Cockrell School of Engineering at the University of Texas at Austin. Dr. Pryor earned his BSME at Southern Methodist University in 1993. He completed his Masters (1999) and PhD (2002) at UT Austin with an emphasis on the modeling, simulation, and operation of redundant manipulators. He has worked for numerous research sponsors including, NASA, DARPA, DOE, INL, LANL, ORNL, Y-12, and many industrial partners. He is a co-founder of the Nuclear Robotics Group and the Drilling \& Rig Automation Group. Both are interdisciplinary research efforts to deploy robotics in hazardous, uncertain environments to perform manufacturing, material handling and other tasks. He is a member of ROS-Industrial, IEEE, ASME, PGE, and ANS.

\bibliographystyle{tfnlm}
\bibliography{main}

\appendix

\section{Additional Figures}
\label{appendix_b}

\begin{figure}[h!]
    \centering
    \includegraphics[scale=0.4]{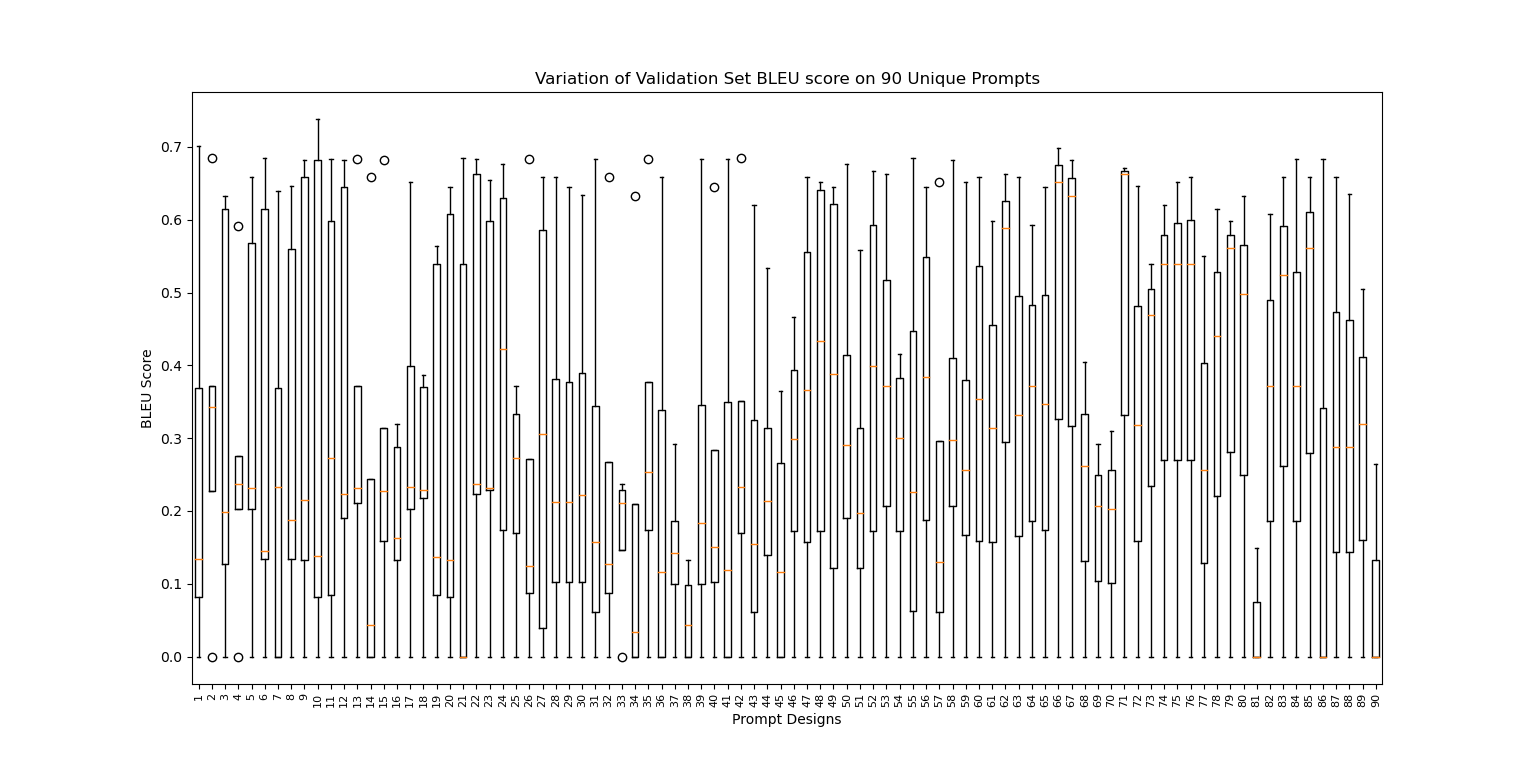}
    \caption{A lack of prompt robustness is shown in this figure.}
    \label{fig:prompting_var}
\end{figure}

\begin{figure}[h!]
    \centering
    \includegraphics[scale=0.4]{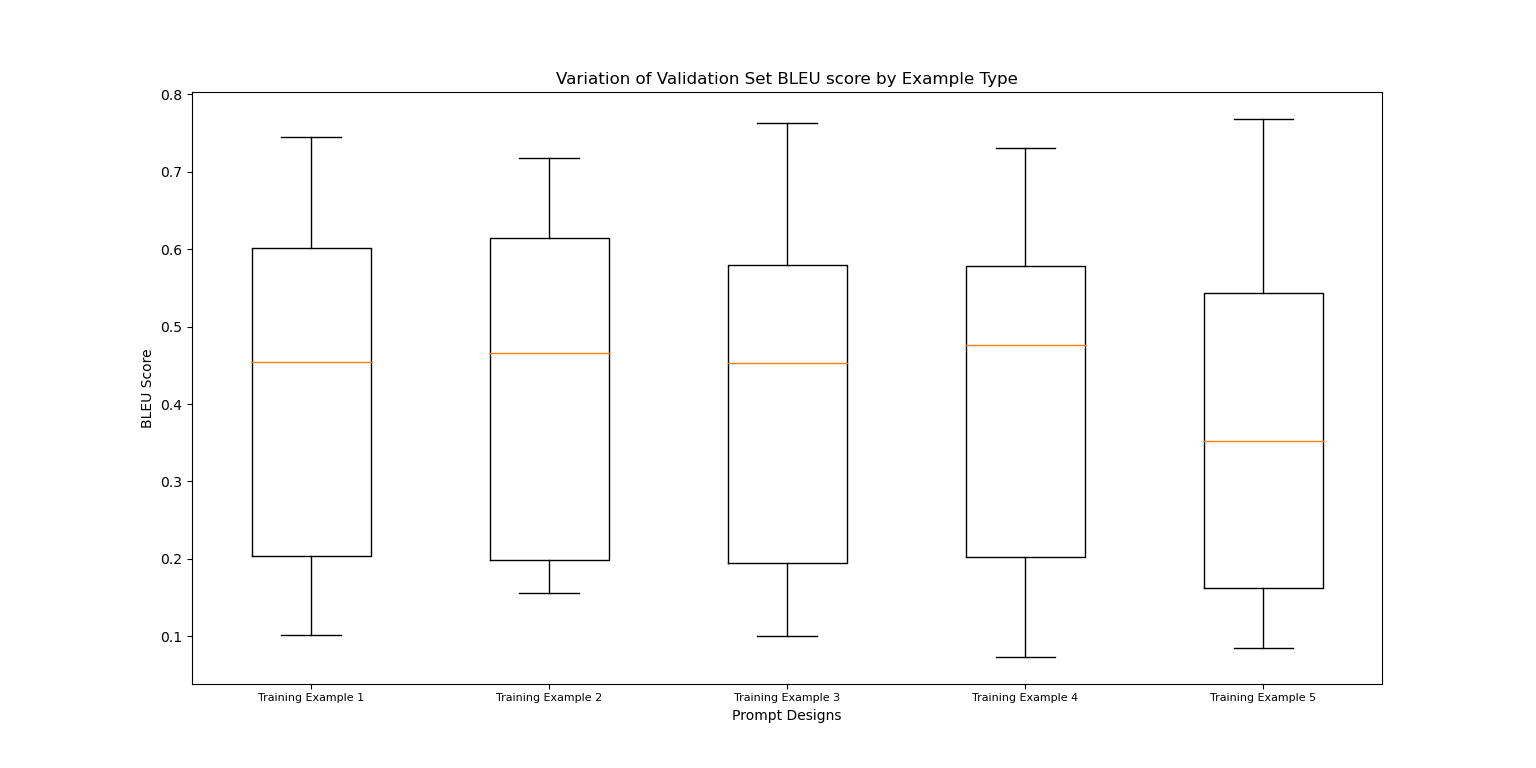}
    \caption{A lack of prompt robustness toward chosen training example is shown in this figure.}
    \label{fig:prompting_ex_importance}
\end{figure}

\begin{figure}[h!]
    \centering
    \includegraphics[scale=0.3]{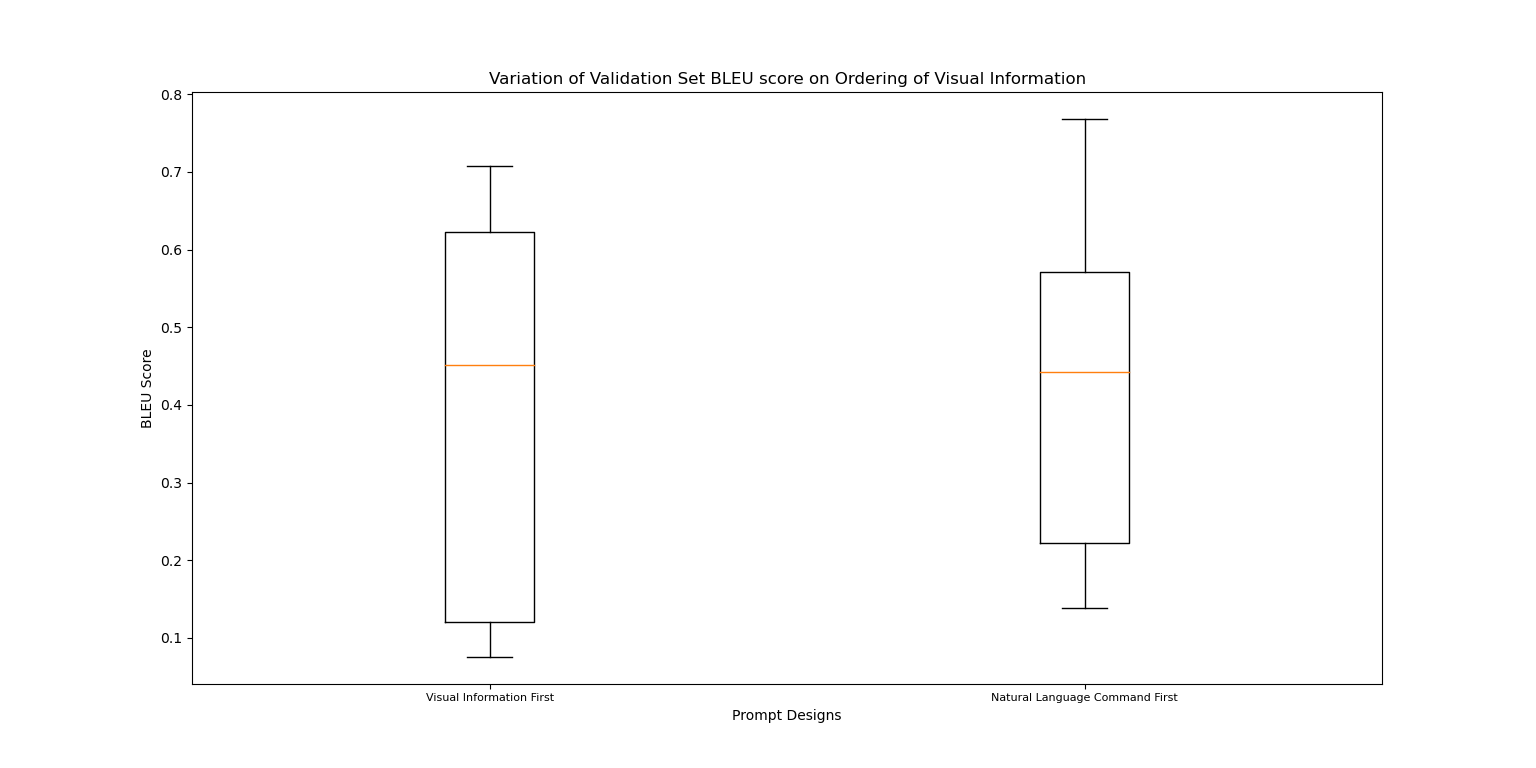}
    \caption{A lack of prompt robustness toward visual versus natural language command cues are shown in this figure.}
    \label{fig:prompt_info_ordering}
\end{figure}

\begin{table}[ht]
\small

\centering
\resizebox{\textwidth}{!}{\begin{tabular}{|p{0.5in}|p{\textwidth-0.5in}|}

\hline
Prompt Type & Prompt Structure\\
\hline
\hline
1  & [x=-9.074; y=-1.89; yaw=2.97] the left side of the same desk + Turn left and approach the left side of the same desk + $\langle$umrf\_label$\rangle$ + ...\\ 
\hline
2  & [x=4.76; y=-6.78; yaw=7.687] the bed + Turn right and face the bed. + $\langle$umrf\_label$\rangle$ + ...\\ 
\hline
3  & [x=-9.15; y=4.316; yaw=2.168] the wall [x=1.26; y=7.61; yaw=-0.214] the table + Turn right and walk to the wall then turn left and walk to the table. +  $\langle$umrf\_label$\rangle$ + ... \\ 
\hline
4  & [x=-6.74; y=-4.67; yaw=3.086] the right side of the wooden desk + Walk over to the right side of the wooden desk. + $\langle$umrf\_label$\rangle$ + ... \\ 
\hline
5  & [x=1.12; y=-1.749; yaw=6.01] the middle of the side of the bed [x=-7.14; y=-3.14; yaw=3.14] the bed + Turn right and take a small step forward then turn right and walk until you're even with the middle of the side of the bed then when you are turn right and walk to the bed. + $\langle$umrf\_label$\rangle$ + ... \\ 
\hline

\end{tabular}
}
\caption{\label{tab: example_types} Prompt structure of example types.
}
\end{table}

\begin{table}[ht]
\small

\centering
\resizebox{\textwidth}{!}{\begin{tabular}{|p{1.5in}|p{3.0in}|p{2.0in}|}

\hline
Prompt Type & Prompt Structure & Average BLEU Score\\
\hline
\hline
61  & Type 4 V + Type 5 L & 0.588\\ 
\hline
65  & Type 4 L + Type 2 L & 0.652\\ 
\hline
66  & Type 4 L + Type 2 V & 0.632\\ 
\hline
70  & Type 5 L + Type 4 V & 0.662\\ 
\hline
73  & Type 5 V + Type 1 V & 0.538\\ 
\hline
74  & Type 5 V + Type 2 L & 0.538\\ 
\hline
75  & Type 5 V + Type 2 V & 0.538\\ 
\hline
78  & Type 5 V + Type 4 L & 0.561\\ 
\hline
82  & Type 5 L + Type 1 V & 0.524\\ 
\hline
84  & Type 5 L + Type 2 V & 0.560\\ 
\hline

\end{tabular}
}
\caption{\label{tab: top_10_struct} Prompt structure of the top 10 best generalizing prompts. The V versus L distinction indicates whether the visual information or the natural language command came first in the example.
}
\end{table}

\end{document}